\documentclass[10pt,twocolumn,letterpaper]{article}
\usepackage{wacv}
\usepackage{times}
\usepackage{epsfig}
\usepackage{graphicx}
\usepackage{amsmath}
\usepackage{amssymb}
\usepackage{multirow}
\usepackage{diagbox}
\usepackage{animate}
\usepackage{mathrsfs}
\usepackage[table,xcdraw]{xcolor}
\usepackage{subfig}
\usepackage{slashbox}
\usepackage{booktabs} 
\usepackage[pagebackref=true,breaklinks=true,letterpaper=true,colorlinks,bookmarks=false]{hyperref}
\wacvfinalcopy 
\def\wacvPaperID{45} 

\ifwacvfinal\fi
\begin{document}
%
\title{Action Quality Assessment Across Multiple Actions}
%
\author{Paritosh Parmar \hspace{2cm} Brendan Tran Morris \\
University of Nevada, Las Vegas\\
{\tt\small parmap1@unlv.nevada.edu, brendan.morris@unlv.edu}
}
%
\maketitle
\ifwacvfinal\thispagestyle{empty}\fi
\begin{figure*}[!t]   
    \centering  
    \captionsetup[subfigure]{labelformat=empty}
    \subfloat[]{\raisebox{0.05in}{\rotatebox{90}{Diving}}}
\animategraphics[loop,autoplay,poster=17,height=0.1\textwidth]{3}{videos/d1/image_}{0001}{0030}
\animategraphics[loop,autoplay,poster=17,height=0.1\textwidth]{3}{videos/d2/image_}{0001}{0030}
\animategraphics[loop,autoplay,poster=17,height=0.1\textwidth]{3}{videos/d3/image_}{0001}{0030}
\animategraphics[loop,autoplay,poster=17,height=0.1\textwidth]{3}{videos/d4/image_}{0001}{0030}
\animategraphics[loop,autoplay,poster=17,height=0.1\textwidth]{3}{videos/d5/image_}{0001}{0030}
\animategraphics[loop,autoplay,poster=17,height=0.1\textwidth]{3}{videos/d6/image_}{0001}{0030}
\animategraphics[loop,autoplay,poster=17,height=0.1\textwidth]{3}{videos/d7/image_}{0001}{0030} \\ \vspace{-1\baselineskip}
    \subfloat[]{ \rotatebox{90}{Gymvault}}
\animategraphics[loop,autoplay,poster=17,height=0.1\textwidth]{3}{videos/g1/image_}{0001}{0030}
\animategraphics[loop,autoplay,poster=17,height=0.1\textwidth]{3}{videos/g2/image_}{0001}{0030}
\animategraphics[loop,autoplay,poster=17,height=0.1\textwidth]{3}{videos/g3/image_}{0001}{0030}
\animategraphics[loop,autoplay,poster=17,height=0.1\textwidth]{3}{videos/g4/image_}{0001}{0030}
\animategraphics[loop,autoplay,poster=17,height=0.1\textwidth]{3}{videos/g5/image_}{0001}{0030}
\animategraphics[loop,autoplay,poster=17,height=0.1\textwidth]{3}{videos/g6/image_}{0001}{0030}
\animategraphics[loop,autoplay,poster=17,height=0.1\textwidth]{3}{videos/g7/image_}{0001}{0030}\\ \vspace{-1\baselineskip}
    \subfloat[]{ \rotatebox{90}{   Skiing}}
\animategraphics[loop,autoplay,poster=17,height=0.1\textwidth]{3}{videos/sk1/image_}{0001}{0030}
\animategraphics[loop,autoplay,poster=17,height=0.1\textwidth]{3}{videos/sk2/image_}{0001}{0030}
\animategraphics[loop,autoplay,poster=17,height=0.1\textwidth]{3}{videos/sk3/image_}{0001}{0030}
\animategraphics[loop,autoplay,poster=17,height=0.1\textwidth]{3}{videos/sk4/image_}{0001}{0030}
\animategraphics[loop,autoplay,poster=17,height=0.1\textwidth]{3}{videos/sk5/image_}{0001}{0030}
\animategraphics[loop,autoplay,poster=17,height=0.1\textwidth]{3}{videos/sk6/image_}{0001}{0030}
\animategraphics[loop,autoplay,poster=17,height=0.1\textwidth]{3}{videos/sk7/image_}{0001}{0030}\\ \vspace{-1.5\baselineskip}
    \subfloat[]{ \rotatebox{90}{Snowboard}}
\animategraphics[loop,autoplay,poster=17,height=0.1001\textwidth]{3}{videos/sn1/image_}{0001}{0030}
\animategraphics[loop,autoplay,poster=17,height=0.1001\textwidth]{3}{videos/sn2/image_}{0001}{0030}
\animategraphics[loop,autoplay,poster=17,height=0.1001\textwidth]{3}{videos/sn3/image_}{0001}{0030}
\animategraphics[loop,autoplay,poster=17,height=0.1001\textwidth]{3}{videos/sn4/image_}{0001}{0030}
\animategraphics[loop,autoplay,poster=17,height=0.1001\textwidth]{3}{videos/sn5/image_}{0001}{0030}
\animategraphics[loop,autoplay,poster=17,height=0.1001\textwidth]{3}{videos/sn6/image_}{0001}{0030}
\animategraphics[loop,autoplay,poster=17,height=0.1001\textwidth]{3}{videos/sn7/image_}{0001}{0030}\\ \vspace{-1.2\baselineskip}
    \subfloat[]{ \rotatebox{90}{S.Dive 3m}}
\animategraphics[loop,autoplay,poster=17,height=0.1\textwidth]{3}{videos/ssd1/image_}{0001}{0030}
\animategraphics[loop,autoplay,poster=17,height=0.1\textwidth]{3}{videos/ssd2/image_}{0001}{0030}
\animategraphics[loop,autoplay,poster=17,height=0.1\textwidth]{3}{videos/ssd3/image_}{0001}{0030}
\animategraphics[loop,autoplay,poster=17,height=0.1\textwidth]{3}{videos/ssd4/image_}{0001}{0030}
\animategraphics[loop,autoplay,poster=17,height=0.1\textwidth]{3}{videos/ssd5/image_}{0001}{0030}
\animategraphics[loop,autoplay,poster=17,height=0.1\textwidth]{3}{videos/ssd6/image_}{0001}{0030}
\animategraphics[loop,autoplay,poster=17,height=0.1\textwidth]{3}{videos/ssd7/image_}{0001}{0030}\\ \vspace{-1.5\baselineskip}
    \subfloat[]{ \rotatebox{90}{S.Dive 10m}}
\animategraphics[loop,autoplay,poster=17,height=0.1\textwidth]{3}{videos/sp1/image_}{0001}{0030}
\animategraphics[loop,autoplay,poster=17,height=0.1\textwidth]{3}{videos/sp2/image_}{0001}{0030}
\animategraphics[loop,autoplay,poster=17,height=0.1\textwidth]{3}{videos/sp3/image_}{0001}{0030}
\animategraphics[loop,autoplay,poster=17,height=0.1\textwidth]{3}{videos/sp4/image_}{0001}{0030}
\animategraphics[loop,autoplay,poster=17,height=0.1\textwidth]{3}{videos/sp5/image_}{0001}{0030}
\animategraphics[loop,autoplay,poster=17,height=0.1\textwidth]{3}{videos/sp6/image_}{0001}{0030}
\animategraphics[loop,autoplay,poster=17,height=0.1\textwidth]{3}{videos/sp7/image_}{0001}{0030}\\ \vspace{-1.5\baselineskip}
    \subfloat[]{ \rotatebox{90}{Trampoline}}
\animategraphics[loop,autoplay,poster=17,height=0.1\textwidth]{3}{videos/t1/image_}{0001}{0030}
\animategraphics[loop,autoplay,poster=17,height=0.1\textwidth]{3}{videos/t2/image_}{0001}{0030}
\animategraphics[loop,autoplay,poster=17,height=0.1\textwidth]{3}{videos/t3/image_}{0001}{0030}
\animategraphics[loop,autoplay,poster=17,height=0.1\textwidth]{3}{videos/t4/image_}{0001}{0030}
\animategraphics[loop,autoplay,poster=17,height=0.1\textwidth]{3}{videos/t5/image_}{0001}{0030}
\animategraphics[loop,autoplay,poster=17,height=0.1\textwidth]{3}{videos/t6/image_}{0001}{0030}
\animategraphics[loop,autoplay,poster=17,height=0.1\textwidth]{3}{videos/t7/image_}{0001}{0030}\\ \vspace{-1\baselineskip}
\caption{\rule{0pt}{3ex}\textbf{Preview of our dataset.} \textit{To see the videos play, please download the manuscript and view in an Adobe Reader.}}
\label{fig:1}
\end{figure*}
\begin{abstract}
Can learning to measure the quality of an action help in measuring the quality of other actions? If so, can consolidated samples from multiple actions help improve the performance of current approaches? In this paper, we carry out experiments to see if knowledge transfer is possible in the action quality assessment (AQA) setting. Experiments are carried out on our newly released AQA dataset (\url{http://rtis.oit.unlv.edu/datasets.html}) consisting of 1106 action samples from seven actions with quality as measured by expert human judges. Our experimental results show that there is utility in learning a single model across multiple actions.
\end{abstract}
%
\section{Introduction}
Action quality assessment (AQA) is the process of quantifying \emph{how well} an action was performed or computing a score representing the execution quality of an action. Automatic AQA, in particular, can be an important component of many practical applications. For \eg, AQA can be employed in low-cost at-home physiotherapy to manage diseases like cerebral palsy \cite{embc16, zhang17}.  High quality AQA can be incorporated into medical training to assess the surgical skills of a student \cite{zia, doughty}. AQA systems can be used to mimic Olympic judges during sports training \cite{pirsia, parmar, venkat, zia}; or provide a second opinion in light of recent judging scandals \cite{olyscand_1, olyscand_2}.  Despite having numerous potential applications, automatic AQA and skill assessment have received little attention in literature. Consequently, there is a dearth of available datasets. 

Current AQA systems/frameworks \cite{pirsia, parmar, venkat, doughty} are concerned with measuring the quality of a single action.  As such, the models are trained on examples of that particular action.  However, existing AQA datasets are typically very small, consisting of only a few samples since the collection of training data requires the use of a domain expert (significant effort by comparison with action recognition).  Due to this data limit, current AQA approaches may not be reaching their full performance potential.  

Pre-training helps with almost every computer vision task. For \eg, object detection \cite{girshick2016region} is aided by pre-training on ImageNet \cite{imagenet}, performance of the C3D network \cite{c3d} was boosted by pre-training on I380K \cite{c3d} and then finetuning Sports-1M \cite{sports1m}, UCF-101 \cite{ucf101} action classification performance was enhanced when the network is pre-trained on Sports-1M \cite{sports1m}, etc. Transfer learning is particularly helpful when target datasets are small. Inspired by several instances where pre-training has helped, we pose the following questions: are there common action quality elements (see Sec. \ref{common_aq_elements}) among different actions? If so, would it be helpful to train/pre-train a model across various actions (instead of following current approach of training on one particular action)? Can a model trained on various actions measure the quality of an unseen action? 

The remaining of the paper is as follows. First, related work in AQA is reviewed in Sec. \ref{sec:related}. Then, a new AQA dataset (the largest to date) is introduced, and notion of common action quality elements is discussed in Sec.  \ref{sec_dataset}. Following which, we answer previously stated questions and introduce a different approach of using all-action model, a model that learns to measure the quality of multiple actions. In Sec. \ref{exps}, we discuss the experiments and results, which provide us the evidence that there is a utility in learning an all-action model over single-action model.
%
%
\section{Related work}
\label{sec:related}
The first major work in the area of AQA was by Pirsiavash \etal \cite{pirsia}, which used pose (of divers and skaters) as features and learned a SVR to map those features to a quality score. By focusing on Olympic sports, they were able to use professionals to provide an objective measure of quality through the judged score. Venkataraman \etal \cite{venkat} extended this work by modifying the feature representation using concatenated approximate entropy of poses rather than DCT of pose.  This better encoding resulted in a 4\% improvement in performance.  Inspired by the success of deep-learning on the task of action recognition, Parmar and Morris \cite{parmar} hypothesized and verified that spatiotemporal features from C3D\cite{c3d} capture the quality aspect of the actions well.  Furthermore, they compared different video feature aggregation schemes, such as averaging clip-level features, and LSTM, to get sample-level features; and regression schemes, SVR, and fully connected layer. 

A unique concept from AQA is feedback for performance improvement.  Pirsiavash \etal \cite{pirsia} were able to suggest minor pose tweaks would could have been adopted in order to improve scores.  The LSTM-based approaches by Parmar and Morris \cite{parmar} did not offer such detailed feedback, but could identify action segments (video clips) where an error might have been made.  These types of error detection mechanisms are important for coaching and faster analysis of athlete performance.  

Steering away from sports, Zia \etal \cite{zia} proposed an approach to measure the surgical skills in an automated fashion. In this work, a classifier is trained on frequency-domain transformed spatiotemporal interest points.  Doughty \etal \cite{doughty} also address the problem of surgical skill assessment as well as the use of chopsticks, dough-rolling, and drawing.  Outside of surgery task, quality was assessed not with a score but based on pairwise comparison of skill in a ``which is better'' framework.  While other quality or skills assessment works do exist, they tend to measure specific characteristics of the task which are not easily generalizable. An example is the basketball skills assessment work by Bertasius \etal \cite{baller}, wherein they present an approach to measure a person's basketball skills from analyzing the video captured by the camera worn by the person. Firstly, atomic basketball events are detected, which are then passed through a Gaussian mixtures to obtain a features which would be representative of player's skills. A skill assessment model is finally learnt on these features to assess basketball skills. Also, it should be noted that skills assessed in \cite{baller} are subjective to evaluator's preference, since annotation was based on a single basketball coach's assessment. In contrast, our approach would not be biased towards any one judge, since our annotated scores are averages of scores from individual judges (usually 5-7 judges).

All of these works use one model each for every action class. Single-acion models or action-specific models don't exploit the fact that there are some common, shared action quality concepts/elements among actions (refer to Sec. \ref{common_aq_elements}). Transfer learning is not new and has been used effectively in the literature\cite{pratt1993discriminability, torrey2010transfer, luo2017label, misra2016cross, torralba2007sharing, yosinski2014transferable, Zamir_2018_CVPR}. In a recent work by Abramovich and Pensky \cite{abramovich}, it was seen that having many classes may be a blessing and not a curse for a classification purposes suggesting that similar result could be true for AQA. In this work, we evaluate to see if there are shared action quality concepts, and if we can exploit them by learning a single model on datapoints from all actions. 
%
\section{AQA-7 Dataset}
\label{sec_dataset}
\begin{table*}[t]
\renewcommand*{\arraystretch}{1.1}
\small
\centering
\begin{tabular}{l|c c c c c}
\toprule

\textbf{Sport}                       & \textbf{Avg. Seq. Len.} & \textbf{\# Samples} & \textbf{Score Range} & \multicolumn{1}{l}{\textbf{\# Participants}} & \multicolumn{1}{l}{\textbf{View Variation}} \\ \midrule
Single Diving 10m platform  & 97                     & 370                     & 21.60 - 102.60       & 1                                           & negligible                                   \\ 
Gymnastic vault             & 87                     & 176                     & 12.30 - 16.87        & 1                                           & large                                        \\ 
Big Air Skiing              & 132                    & 175                     & 8 - 50               & 1                                           & large                                        \\  
Big Air Snowboarding        & 122                    & 206                     & 8 - 50               & 1                                           & large                                        \\ 
Sync. Diving 3m springboard & 156                    & 88                      & 46.20 - 104.88       & 2                                           & negligible                                   \\  
Sync. Diving 10m platform   & 105                    & 91                      & 49.80 - 99.36        & 2                                           & negligible                                   \\
Trampoline                  & 634                    & 83                      & 6.72 - 62.99         & 1                                           & small                                        \\ \toprule
\end{tabular}
\caption{\textbf{Characteristics of \textsc{AQA-7} dataset.}}
\label{table:1}
\end{table*} 
To the best of our knowledge, there are only two publicly available AQA datasets \cite{pirsia, parmar} and they have limited number of samples for each individual activity. This is partly due to the extra effort required to collect samples as compared to an action recognition dataset. In case of action recognition dataset compilation, an annotator might go to video hosting website such as YouTube and run a search query on names of actions. While in compiling an AQA dataset, annotator has to mark starting and ending frames of a sample, and note down execution score, difficulty level, final score, etc. Additionally, field experts are required to evaluate and score AQA data samples which can be quite difficult in many domains. These factors limit dataset size. 

Furthermore, for action recognition, many more meaningful short samples can be produced from one original longer video. By meaningful sample, we mean that, even if you clip an action recognition video, you would still have the action class concept captured -- it has been shown that action classification task maybe performed using as little as a single frame \cite{sports1m}. Unlike action recognition, in case of AQA, a full action sequence must be examined because an error in execution can be made at any time during the action sequence.  Leaving out frames where the error occurred would result in predicted scores that are not indicative of the true quality.  This limits data augmentation tricks for more effective sample utilization. 

To fill this data void, we introduce Action Quality Assessment 7 (AQA-7) dataset (Fig. \ref{fig:1}), comprising samples from seven actions: \{singles diving-10m platform, gymnastic vault, big air skiing, big air snowboarding, synchronous diving-3m springboard, synchronous diving-10m platform, and trampoline\} captured during Summer and Winter Olympics. Scoring rules change from time to time; however, we made sure that rules were same for the samples that we collected. A summary of the AQA-7 dataset characteristics are given in Table \ref{table:1}. \\ 
\subsection{Description of Actions}
\textbf{Diving: }
Individual diving has been a Summer Olympic sport since the early 1900s and synchronized diving was introduced in 2000.  Divers perform acrobatic stunts such as somersaults and twists either from a platform or a springboard.  The score is based on the difficulty of and the quality of execution of the dive.  AQA-7 contains three different events -- the 10m platform (Diving in first row of Fig. \ref{fig:1}), synchronized 10m platform (Sync Dive 10m in sixth row of Fig. \ref{fig:1}), and 3m springboard (Sync Dive 3m in fifth row of Fig. \ref{fig:1}).  The three diving events have negligible view variation as they are all shot from a consistent angle for all samples.  The synchronized dives have two athletes participating and their synchronization, which is given more importance than execution, is also a measure of quality of the action. The singles diving-10m platform class is an extension of the 159 samples originally from Pirsiavash et al. \cite{pirsia} to 370 dives. \\
\textbf{Gymnastic vault: }
Gymvault is a gymnastic event from the Summer Olympics where the athlete runs down a runway and uses a springboard to launch themselves over the vault and perform aerial tricks.  The gymvault dataset was collected from various international events in addition to Olympics.  Like diving, the quality of gymvault is assessed based on the difficulty and the quality of execution.  Examples tend to have large view variation both over the course of a single vault attempt (running, planting on vault, spinning in air, and landing) as well as due to differences broadcast camera placement at different events (see second row of Fig. \ref{fig:1}).  \\
\textbf{Big Air: }
The Ski (BigSki) and Snowboard (BigSnow) Big Air events were new events in the 2018 Winter Olympics.  Samples were obtained from previous X-Games since it is the premiere venue for the event.  Big Air has significant view variation between events and within an example due to camera location (rows 3 and 4 of Fig. \ref{fig:1}).  Scoring is based on four components -- difficulty, execution, amplitude, and progression and landing -- a more complicated formula than the other actions especially since there is a qualitative component of pushing Big Air forward by doing tricks that nobody else is doing.  Note that for Snowboard Big Air the pixel coverage size of the person is similar to that of the snowboard.\\
\textbf{Trampoline: }
The Trampoline event was added to the Olympics in 2000.  It is judged based on difficulty, execution, and time of (in air) flight.  Like for diving, the camera is in a consistent side-view position but also moves up and down to follow the athlete for minimal view variation.  Unlike all the other sports which consist of a single ``action'' performed in less than five seconds, a trampoline sample will contain multiple (10) action phases, where the athlete twists and turns, over its 20 seconds.  
\subsection{Common Action Quality Elements}
\label{common_aq_elements}
The AQA-7 sports have similar action elements which include flipping and twisting (Fig. \ref{fig:1}).  As such, the quality of actions are assessed in similar fashion.  For \eg in Diving and Gymvault, judges expect athletes to have legs perfectly straight in pike position (execution quality aspect) and degree of difficulty is directly proportional to the number of twists and somersaults.  Similarly, in ski and snowboard Big Air events, the difficulty is related to the number of vertical and horizontal spins.  In all sports there is a expectation of a perfect landing with high impact on final score (an entry with minimal splash or `ripping' is the diving equivalent).  

The reason behind these similarities is that having to complete more number of somersaults, twists, or spins (difficulty aspect) while keeping legs straight, and having the body in tight tuck or pike position (execution quality aspect) in the limited time from take-off to landing makes it harder to achieve and, therefore, worthy of more points from judges (equivalently, higher quality). The final score is a function of execution quality aspect and the degree of difficulty.  In some cases, like Diving, it will be the product function, in another case, like Gymault, it will be the summation, while other actions (BigAir events) may use a more holistic combination approach. Given the similarities in actions, it is believed that knowledge of what aspects to value in one action can help in measuring the quality in another. 
%
%
\section{Our Approach}
\begin{figure}[]
\includegraphics[width=\linewidth]{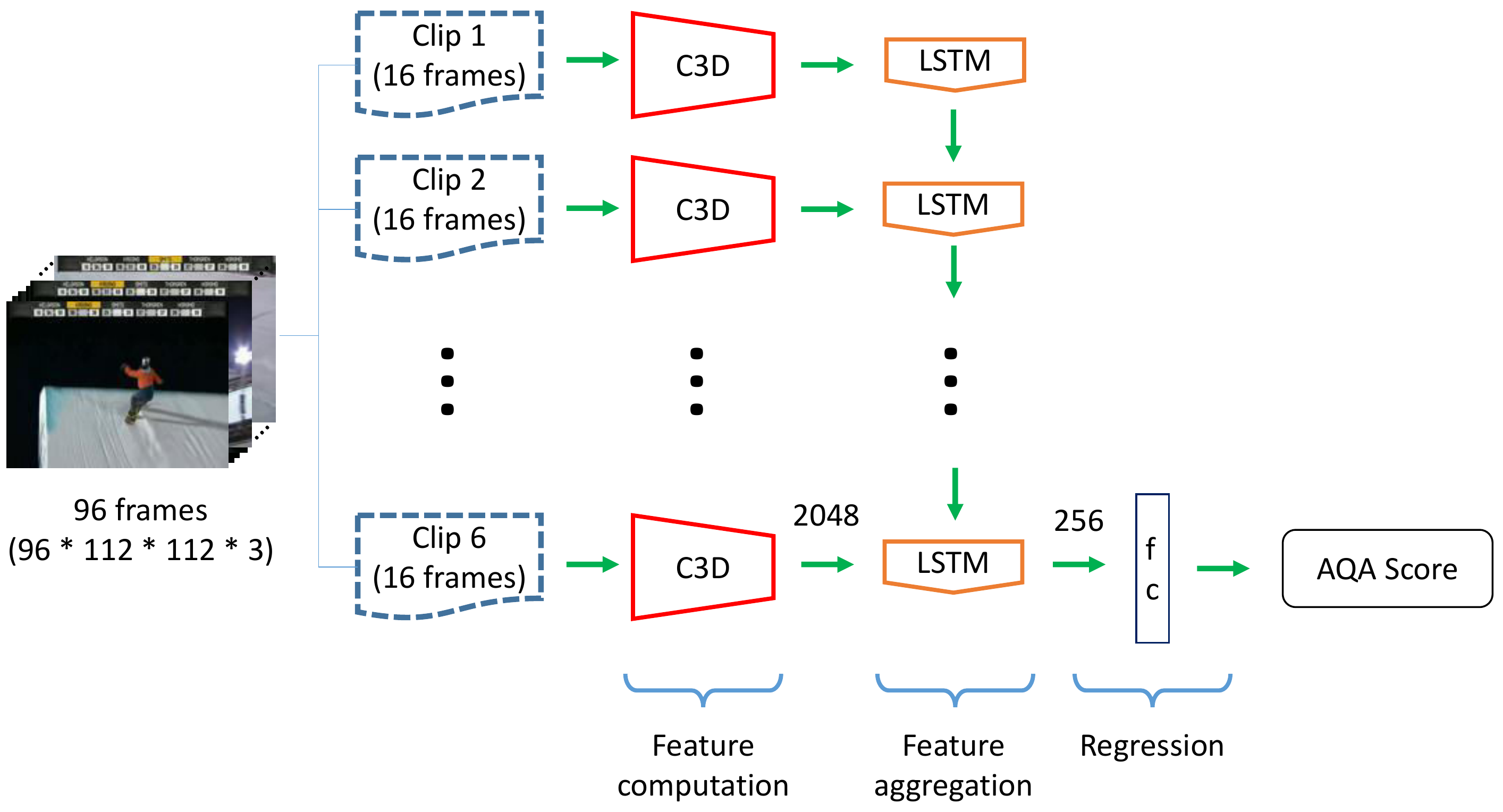}
\caption{\textbf{\textsc{C3D-LSTM} network}}
\label{fig_c3d_lstm}
\end{figure}

Although, in general, action quality may be highly dependent on the action, we hypothesize that actions do have commonalities (Sec. \ref{common_aq_elements}) that can be exploited despite individual differences such as judging criteria. However, it may be that each action is unique and do not have common quality elements and our hypothesis is wrong, and hence it would not be possible to share knowledge or learn a consistent model that can measure quality across multiple actions.

In order to test if our hypothesis is correct, and see if we can transfer action quality knowledge across actions, we propose to use the C3D-LSTM framework \cite{parmar}, illustrated in Fig. \ref{fig_c3d_lstm}. This network consists of a smaller version of C3D network \cite{c3d} followed by a single 256-dimensional LSTM \cite{lstm} layer and a fully-connected (\texttt{fc}) layer, which outputs the final AQA score. A video is processed in clips of 16 frames to generate C3D features at the \texttt{fc6} layer which is connected to the LSTM layer for temporal feature aggregation. The C3D network is kept frozen during training so only the LSTM and final \texttt{fc} layer parameters are tuned. Euclidean distance between the predicted scores and true scores is used as the loss function to be minimized. The main difference in this work is rather than building an individual model for each action (we refer to these as action-specific or single-action models), a single model is learned by training with samples from all/multiple actions (we refer to our model as all-action or multi-action models).

To answer our fundamental questions regarding quality elements shared across actions, three different experiments are designed: 1) to check if it is possible to learn an all-action, and if so, compare the performance of our proposed all-action model against action-specific model (Sec. \ref{exp_1}), 2) to evaluate how well can an all-action model quantify the quality of unseen action classes (Sec. \ref{exp_2}), and 3) to evaluate the generalization of all-action model to novel action classes (Sec. \ref{exp_3}).  
\section{Experiments}
\label{exps}
\begin{table*}[t]
\centering
\small
\begin{tabular}{l|cccccc|c}
\cmidrule[\heavyrulewidth]{2-8}
\multicolumn{1}{l|}{}                                                      & \textbf{Diving} & \textbf{Gymvault} & \textbf{Skiing} & \textbf{\begin{tabular}[c]{@{}c@{}}Snowb-\\ oarding\end{tabular}} & \textbf{\begin{tabular}[c]{@{}c@{}}Sync.\\ Dive 3m\end{tabular}} & \textbf{\begin{tabular}[c]{@{}c@{}}Sync.\\ Dive 10m\end{tabular}} & \textbf{\begin{tabular}[c]{@{}c@{}}Avg.\\ Corr.\end{tabular}} \\ \midrule
Pose+\textsc{DCT} \cite{pirsia}                                                           & 0.5300          & -                 & -               & -                                                                 & -                                                                & -                                                                 & -                                                             \\ 
\begin{tabular}[c]{@{}l@{}}Single-action\\ \textsc{C3D-SVR}\cite{parmar}\end{tabular}    & \textbf{0.7902} & \textbf{0.6824}   & \textbf{0.5209} & 0.4006                                                            & 0.5937                                                           & \textbf{0.9120}                                                   & \textbf{0.6937}                                               \\ \midrule
\begin{tabular}[c]{@{}l@{}}Single-action\\ \textsc{C3D-LSTM}\cite{parmar}\end{tabular}   & 0.6047          & 0.5636            & 0.4593          & \textbf{0.5029}                                                   & 0.7912                                                           & 0.6927                                                            & 0.6165                                                        \\ 
\begin{tabular}[c]{@{}l@{}}Ours All-action\\ \textsc{C3D-LSTM}\end{tabular} & 0.6177          & 0.6746            & 0.4955          & 0.3648                                                            & \textbf{0.8410}                                                  & 0.7343                                                            & 0.6478                                                        \\ \bottomrule
\end{tabular}
\caption{\textbf{All-Action vs. Single-Action models}. Performance evaluation of single-action and all-action models in terms of action-wise and average Spearman's rank correlation (higher is better). First two frameworks simply average features to aggregate them and use SVR as the regression module. The bottom two frameworks use LSTM to aggregate features and use a fully-connected layer as the regression module. Our approach can be directly compared with single-action C3D-LSTM \cite{parmar}, since both have the same architecture. }
\label{tab_all-action}
\end{table*}

\begin{table*}[t]
\centering
\small
\begin{tabular}{l|rrrrrr|r}
\toprule
\multicolumn{1}{l|}{\diagbox[]{\textbf{Training}\\\textbf{action}\textbf{ class}}{\textbf{Unseen}\textbf{ action}\\\textbf{class}}} & \multicolumn{1}{c}{\textbf{Diving}} & \multicolumn{1}{c}{\textbf{\begin{tabular}[c]{@{}c@{}}Gym-\\vault\end{tabular}}} & \multicolumn{1}{c}{\textbf{Skiing}} & \multicolumn{1}{c}{\textbf{\begin{tabular}[c]{@{}c@{}}Snow-\\board\end{tabular}}} & \multicolumn{1}{c}{\textbf{\begin{tabular}[c]{@{}c@{}}Sync-\\Dive 3m\end{tabular}}} & \textbf{\begin{tabular}[c]{@{}c@{}}Sync-\\Dive 10m\end{tabular}} & \textbf{\begin{tabular}[c]{@{}c@{}}Avg.\\Corr.\end{tabular}} \\ \midrule
\textbf{\begin{tabular}[c]{@{}r@{}}Random Wts./Ini.\end{tabular}} & 0.0590                               & 0.0280                                                                             & -0.0602                              & -0.0703                                                                             & -0.0146                                                                               & -0.0729                                                           & -0.0218                                                      \\ \midrule   
\textbf{Diving}         & \textbf{0.6997}                               & -0.0162                                                                            & 0.0425                               & 0.0172                                                                              & 0.2337                                                                                & 0.0221                                                            & 0.0599                                                       \\ 
\textbf{Gymvault}       & 0.0906                               & \textbf{0.8472}                                                                   & 0.0517                               & 0.0418                                                                              & -0.1642                                                                               & -0.3200                                                           & -0.0600                                                      \\ 
\textbf{Skiing}         & 0.2653                               & -0.1856                                                                            & \textbf{0.6711}                               & 0.1807                                                                              & 0.1195                                                                                & 0.2858                                                            & 0.1331                                                       \\ 
\textbf{Snowboard}      & 0.2115                               & -0.2154                                                                            & 0.3314                               & \textbf{0.6294} & 0.0945                                                                                & 0.1818                                                            & 0.1208                                                       \\ 
\textbf{Sync. Dive 3m}  & 0.1500                               & -0.0066                                                                            & -0.0494                              & -0.1102                                                                             & \textbf{0.8084} & 0.0428                                                            & 0.0053                                                       \\ 
\textbf{Sync. Dive 10m} & 0.0767                               & -0.1842                                                                            & 0.0679                               & 0.0360                                                                              & 0.4374                                                                                & \textbf{0.7397}                                                            & 0.0868                                                       \\ \midrule 
\textbf{Multi-action}                                                      & 0.2258                               & 0.0538                                                                             & 0.0139                               & 0.2259                                                                              & 0.3517                                                                                & 0.3512                                                            & \textbf{0.2037}                                              \\ \bottomrule
\end{tabular}
\caption{\textbf{Zero-shot AQA}. Performance comparison of randomly-initialized model, single-action models (for \eg, first row shows the results of training on diving action measuring the quality of the remaining (unseen) action classes), and multi-action model (all-action model trained on five action classes) on unseen action classes. In multi-action class, the model is trained on five action classes and tested on the remaining action class (column-wise). In single-action model rows, diagonal entries show results of training and testing on the same action. Avg. Corr. shows the result of average (using Fisher's z-score) correlation across all columns.}
\label{tab_zeroshot_aqa}
\end{table*}

\begin{figure*}[!t]
\centering
\subfloat[Diving - 25]{\includegraphics[width = 1.1in]{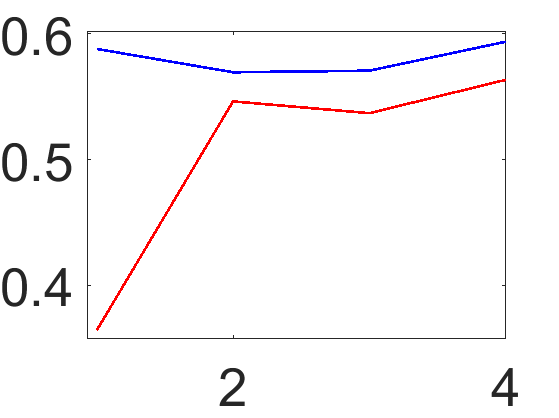}} 
\subfloat[Diving - 75]{\includegraphics[width = 1.1in]{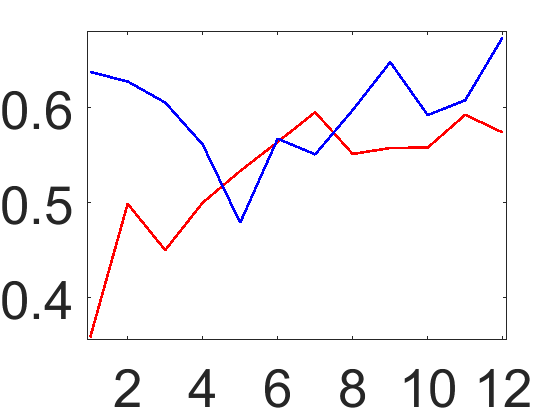}}
\subfloat[Diving - 125]{\includegraphics[width = 1.1in]{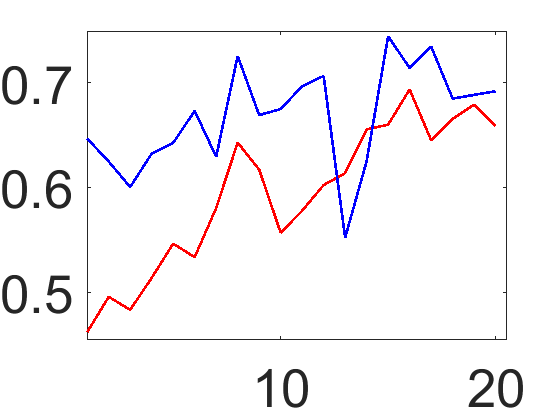}} 
\subfloat[Gymv - 25]{\includegraphics[width = 1.1in]{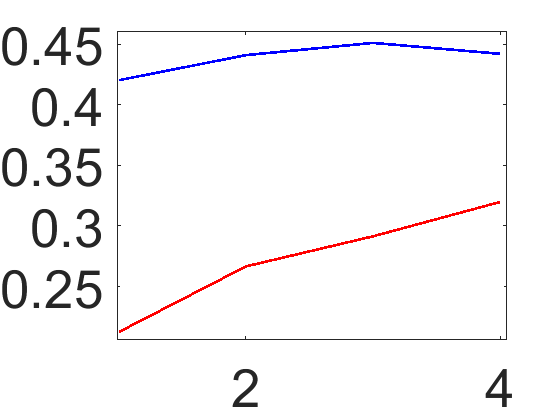}} 
\subfloat[Gymv - 75]{\includegraphics[width = 1.1in]{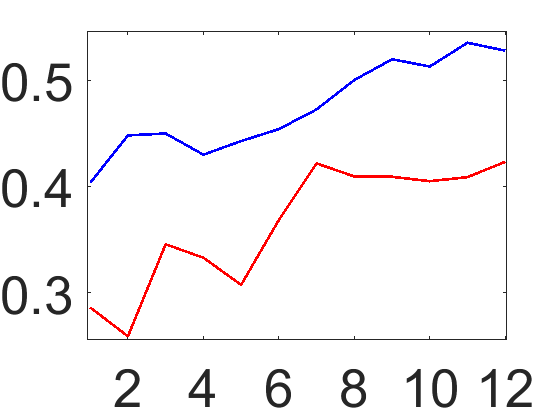}}
\subfloat[Gymv - 125]{\includegraphics[width = 1.1in]{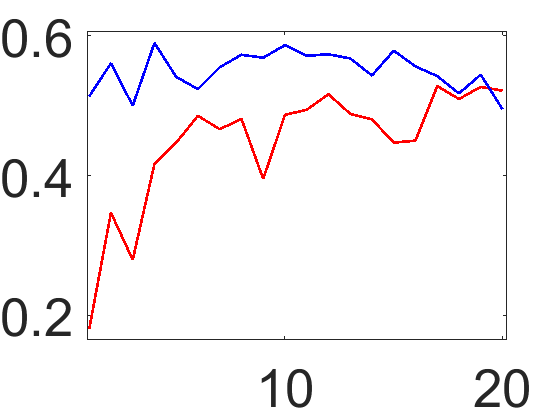}} \\
\subfloat[Ski - 25]{\includegraphics[width = 1.1in]{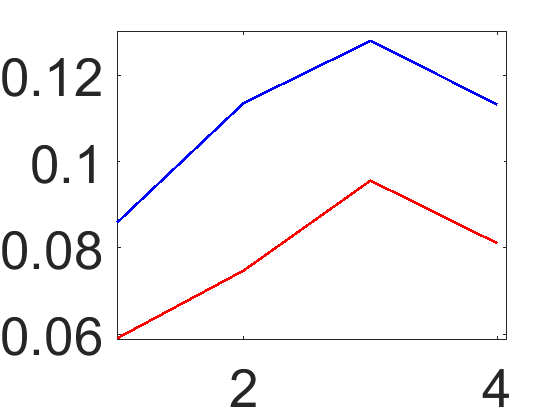}} 
\subfloat[Ski - 75]{\includegraphics[width = 1.1in]{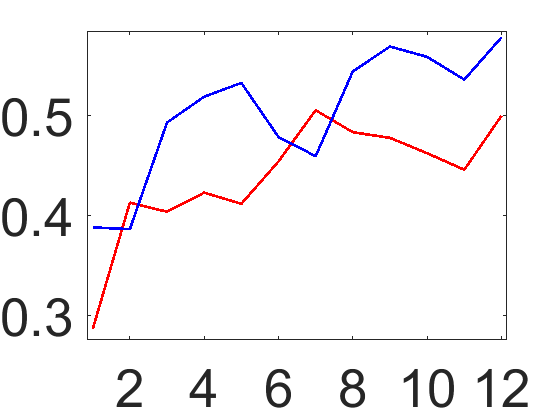}}
\subfloat[Ski - 125]{\includegraphics[width = 1.1in]{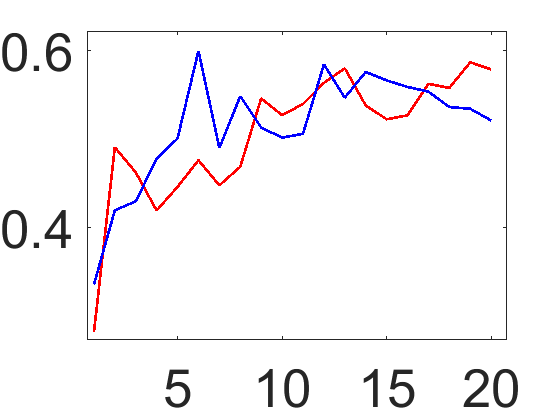}} 
\subfloat[Snowb - 25]{\includegraphics[width = 1.1in]{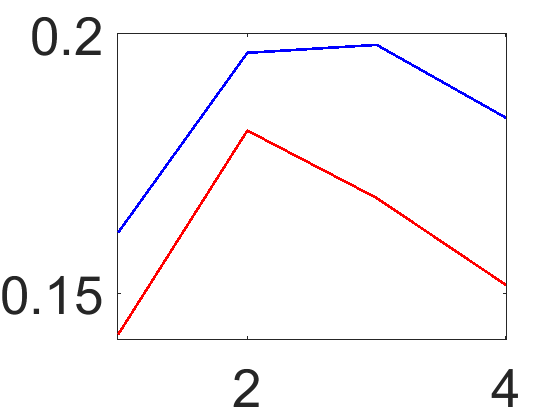}} 
\subfloat[Snowb - 75]{\includegraphics[width = 1.1in]{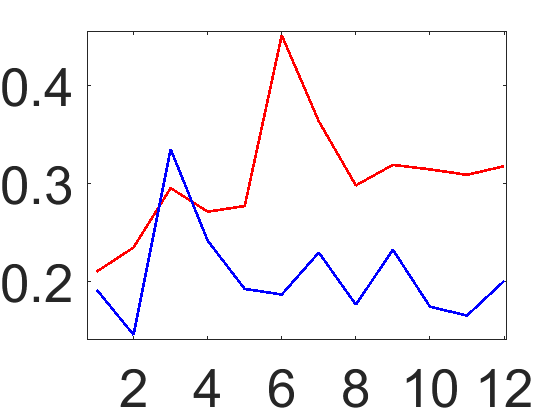}}
\subfloat[Snowb - 125]{\includegraphics[width = 1.1in]{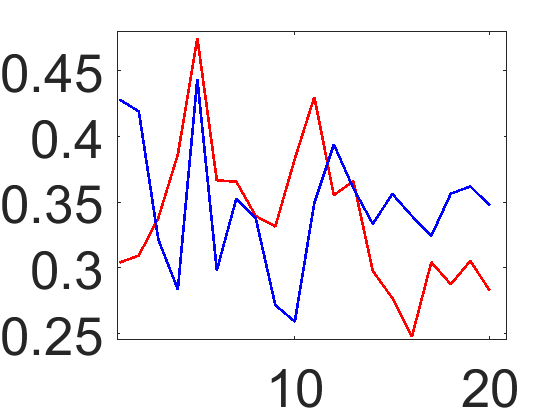}} \\
\subfloat[S.D 3m - 15]{\includegraphics[width = 1.1in]{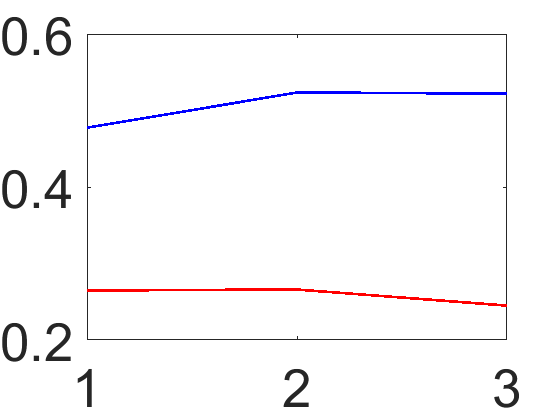}} 
\subfloat[S.D 3m - 25]{\includegraphics[width = 1.1in]{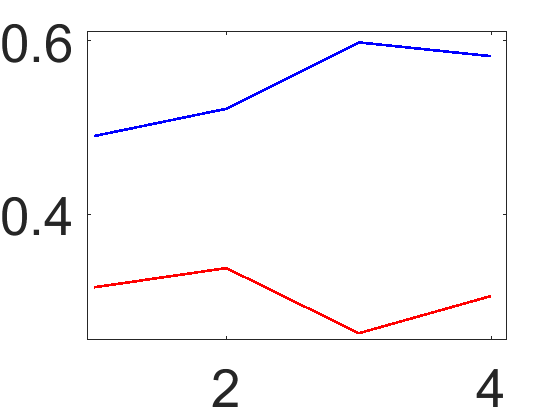}}
\subfloat[S.D 3m - 35]{\includegraphics[width = 1.1in]{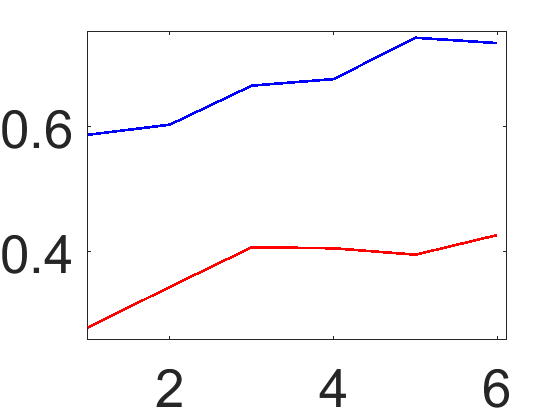}} 
\subfloat[S.D 10m - 15]{\includegraphics[width = 1.1in]{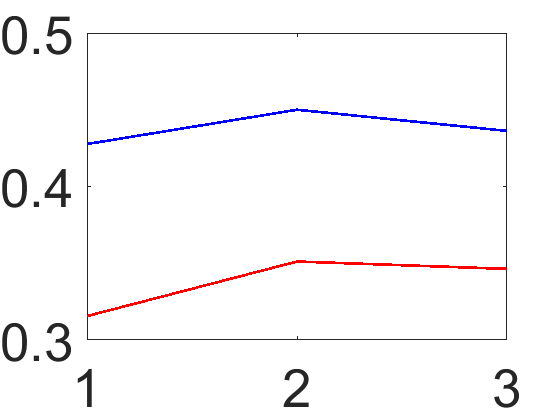}} 
\subfloat[S.D 10m - 25]{\includegraphics[width = 1.1in]{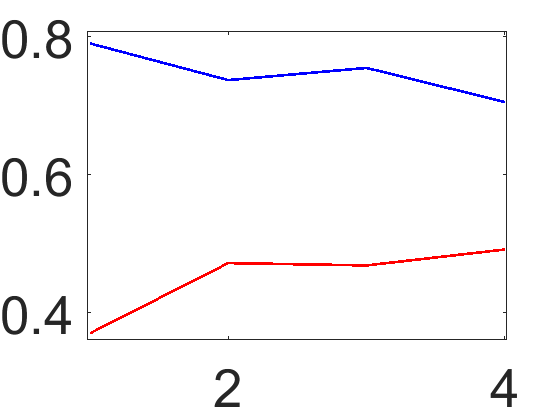}}
\subfloat[S.D 10m - 35]{\includegraphics[width = 1.1in]{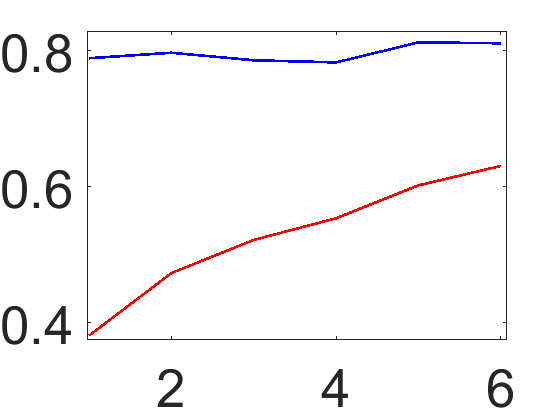}} \\
\caption{\textbf{Finetuning from scratch vs. finetuning from pre-trained multi-action model.} Plot of Spearman's rank correlation against every hundred iterations for different number of training samples. Blue and red curves represent multi-action and randomly initialized models, respectively. The gap in the initial iterations suggest that good initialization of LSTM weights was achieved by training on multiple actions. In most of the cases, multi-action model has better performance than randomly initialized model on test samples throughout all the iterations.}
\label{fig_ft_plots}
\end{figure*}
\begin{table*}[t]
\centering
\small
\begin{tabular}{l|ccccccccc}
\toprule
\textbf{Test action} & \multicolumn{3}{c|}{\textbf{Diving}}                                                                    & \multicolumn{3}{c|}{\textbf{Gymvault}}                                                                  & \multicolumn{3}{c}{\textbf{Skiing}}                                                                    \\ \midrule
\textbf{\# samples}  & \multicolumn{1}{c|}{\textbf{25}} & \multicolumn{1}{c|}{\textbf{75}} & \multicolumn{1}{c|}{\textbf{125}} & \multicolumn{1}{c|}{\textbf{25}} & \multicolumn{1}{c|}{\textbf{75}} & \multicolumn{1}{c|}{\textbf{125}} & \multicolumn{1}{c|}{\textbf{25}} & \multicolumn{1}{c|}{\textbf{75}} & \multicolumn{1}{c}{\textbf{125}} \\ \midrule
\textbf{RI}          & 0.5633                           & 0.5952                           & 0.6935                            & 0.3197                           & 0.4231                           & 0.5278                            & 0.0955                           & 0.5050                           & 0.5862                            \\ 
\textbf{AA}        & \textbf{0.5937}                           & \textbf{0.6742                          } & \textbf{0.7443}                            & \textbf{0.4509                          } & \textbf{0.5350}                           & \textbf{0.5894                           } & \textbf{0.1279}                           & \textbf{0.5778}                           & \textbf{0.5991}                            \\ \bottomrule
\end{tabular}
\\
\rule{0pt}{8ex}
\begin{tabular}{ccccccccc}
\toprule
\multicolumn{3}{c|}{\textbf{Snowboard}}                                                                 & \multicolumn{3}{c|}{\textbf{Sync. Dive 3m}}                                                            & \multicolumn{3}{c}{\textbf{Sync. Dive 10m}}                                                           \\ \midrule
\multicolumn{1}{c|}{\textbf{25}} & \multicolumn{1}{c|}{\textbf{75}} & \multicolumn{1}{c|}{\textbf{125}} & \multicolumn{1}{c|}{\textbf{15}} & \multicolumn{1}{c|}{\textbf{25}} & \multicolumn{1}{c|}{\textbf{35}} & \multicolumn{1}{c|}{\textbf{15}} & \multicolumn{1}{c|}{\textbf{25}} & \multicolumn{1}{c}{\textbf{35}} \\ \midrule
0.1813                            & \textbf{0.4507}                           & \textbf{0.4751}                            & 0.2659                           & 0.3382                           & 0.4268                           & 0.3511                           & 0.4913                           & 0.6305                           \\
\textbf{0.1978}                            & 0.3347                           & 0.4437                            & \textbf{0.5235}                           & \textbf{0.5980}                           & \textbf{0.7429}                           & \textbf{0.4500}                           & \textbf{0.7900}                           & \textbf{0.8123}                         \\ \bottomrule
\end{tabular}

\caption{\textbf{Finetuning from scratch vs. finetuning from pre-trained multi-action model.} Experimental results (Spearman's rank correlation) of finetuning a randomly-initialized (RI) model and an all-action (AA) model pre-trained on five action classes. The numbers represent the best results from all the iterations.}
\label{tab_ri_5vs1}
\end{table*}
Three different experiments are performed to test whether it is advisable to learn a joint action quality model rather than individual model.  To be consistent with existing literature, Spearman's rank correlation (higher is better) is used as the performance metric. When presenting aggregated results, the average Spearman's rank correlation is computed from individual per action correlations using Fisher's z-value as described in \cite{garcia2010tutorial}.

\textbf{Data Preparation:} Since different actions have different ranges for scores, as shown in Table \ref{table:1}, we divide the raw scores of all the actions by the training standard deviation of the corresponding action. At test time, we multiply the predicted scores by the standard deviation of the appropriate action to get the final judged value.  Experiments were only conducted on six of the action classes since they are of similar short length.  All videos are normalized to a fixed size of 103 frames by zero-padding the first frames where needed. Trampoline is excluded from the experiments since the average length of 650 frames is much longer and composed of multiple ``tricks.''  The model is trained using 803 videos, and tested on the remaining 303 videos. During training, temporal data augmentation is used to have six different copies of the same video sample with one frame start difference (effectively 4818 training samples).

\textbf{Implementation details:} Caffe\cite{caffe} was used to implement our model on a Titan-X GPU. The C3D network was pretrained on UCF-101 \cite{ucf101} for 100k iterations with an initial learning rate of 0.001 and annealed by a factor of 10 after every 40k iterations, momentum of 0.9, and a weight decay of 0.005. After pre-training, C3D is frozen and used as a feature extractor.

For the LSTM layer, the ADAM solver \cite{adam} is used with an initial learning rate of 0.001 and annealed by a factor of 2 after every 3K iterations. Optimization was carried out for 20K iterations after the LSTM layer was initialized with Gauassian noise with standard deviation of 0.1. Using a stride of 16 frames, the C3D network ``sees'' a whole action sequence in 6 LSTM steps. We use a batch size of 90 clips (15 full video samples).
\subsection{All-Action vs. Single-Action Models}
\label{exp_1} 
In the first experiment, rather than learning a model using only samples from a single action, our All-action model is learned using diverse and larger dataset with samples from across six actions.  A comparison of the all-action model with other state-of-the-art approaches is given in Table \ref{tab_all-action}.    

Pirsiavash \etal \cite{pirsia} optimized their pose estimator for each action (diving, figure-skating) and therefore cannot be compared with actions other than diving fairly, so we provide their results only on diving. The two other comparisons, C3D-SVR and C3D-LSTM, are the state-of-the-art single-action (action-specific) results from Parmar and Morris \cite{parmar}. Pose+DCT \cite{pirsia} and C3D-SVR \cite{parmar} use SVR as regression module. The baseline for the All-action model is the single-action C3D-LSTM since both use same feature aggregation method (LSTM) and regression module (\texttt{fc} layer). 

It is clear that the use of spatio-temporal features (C3D net, \texttt{fc6} layer activations) improves over Pose+DCT. The proposed all-action model outperforms the single-action model for five of the six actions -- all but for Snowboarding which was less by 0.14.  On average across the actions, the all-action model improves Spearman's rank correlation performance by 0.03 without changing the network and instead leveraging data samples from all actions.  

The 0.6478 correlation value from all-action model is competitive even with action-specific C3D-SVR's 0.6937 (six different C3D-SVR's are used, one for each action).  This is noteworthy since C3D-SVR was the best performing AQA system in \cite{parmar} and, as they mentioned, LSTM aggregation is preferred for its ability to temporally localize action quality drops (error identification). 

Note that our strategy to train a single model using datapoints from all the actions is complementary to the existing approaches and employing it may help improve their performance.
\subsection{Zero-Shot AQA}
\label{exp_2}
One open question for AQA is whether the learned concept of quality can generalize. If action quality concepts are shared among actions, then the knowledge of how to measure the quality of a group of actions, is likely to help in measuring the quality of other, unseen action/s.
\subsubsection{Random-initialization vs. Multi-action pre-training}
While the previous experiment showed that all-action C3D-LSTM model performed better than the single-action counterpart, it did not necessarily indicate that learning one action helped to learn another action.  To address this issue, a zero-shot AQA experiment is designed.  In the zero-shot setting, a model is trained on five actions and tested on the remaining unseen sixth action.  The baseline for comparison is the same network with random LSTM weights (C3D weights remain same for all models). Comparing multi-action model with randomly-initialized weights may not seem very fair, but interestingly, a hierarchy of random, untrained convolution filters have been shown to perform almost as well as learned filter weights in a work by Jarrett \etal \cite{jarrett}. It is assumed that if innate quality concepts were \textit{not} learned, then the resulting multi-action model (an all-action model trained using five action classes) would have similar performance to random initialization.  If instead the multi-action model is much better than random initialization on the zero-shot AQA task then it indicates that there are common quality elements across actions and that there is utility in pre-training a model across multiple actions.

A summary of AQA performance on an unseen action is detailed in Table \ref{tab_zeroshot_aqa}.  With random initialization the AQA system is not able to perform with any reliability, as indicated by Spearman's rank correlation close to zero value.  In contrast, the all-action version shows some positive correlation. We believe that the reason for better performance of the all-action model is that the use of multiple actions provides a good initialization since there are common/shared action quality elements.  In addition, with all-action model there is an advantage of having access to more training videos from which to learn. Although, our all-action works better than random-initialization across four actions, there are two actions - Gymvault, and Skiing - that seem to be providing very weak indication. So, we explore further (refer Sec. \ref{exp_3}) in order to be sure that our proposed all-action actually attains a good initialization, in general.
\subsubsection{Single-Action vs. Multi-Action Transfer}
To further examine knowledge transfer, a sub-experiment is conducted to assess the performance of a single-action model on the unseen action classes. The results are compiled in middle six rows of Table \ref{tab_zeroshot_aqa}.  Note: single-action pre-training is actually a special case of our proposed multi-action pre-training.

When examining the single-action results, note they resemble a confusion matrix with high diagonal elements (matched training and testing data) but with low off-diagonals (mismatched test classes to training class).  The single-action models tend to perform well on very similar actions e.g. Diving on Sync. Dive 3m/10m, or vice-versa, or Snowboarding on Skiing, and vice-versa.  This result is intuitive since similar actions have more shared concepts.  However, non-intuitive relationships are apparent as well.  An example is the Skiing model tested on Diving, which have quite different judging criteria yet still manages significant rank correlation.  While no one single-action model is strong across all classes it does support the idea of shared quality concepts across actions.

On average, the All-action model greatly outperforms any single-action model. With more action classes included in training, more quality concepts/elements are shared and there are more datapoints from which to learn each of the elements of quality resulting in better learning of the concepts.  Also, the ratio of the overall number of quality concepts to the total number of datapoints is reduced due to more sharing of elements. In other words, as we increase our action bank, by including more actions, chances of sharing quality elements with an unseen action increase. 
\subsection{Fine-tuning to a Novel Action Class}
\label{exp_3}
In the final experiment, fine-tuning is used to adjust a pre-trained model for a new unseen action.  The experiment considers the situation where there are limited number of samples in the dataset of a newly considered action to learn with (\ie it is difficult to obtain many labeled samples as might be the case for an obscure action).  

In particular, the all-action model is pre-trained on five actions and fine-tuned on the remaining unseen action with minimal datapoints.  For the more data rich actions (Diving, Gymvault, Skiing, and Snowboarding) \{25, 75, 125\} training samples are used, while for data poor (Sync. Dive 3m/10m) only \{15, 25, 35\} training samples are used for fine-tuning.  Testing is performed on 50 of the remaining samples for all actions.  The settings are the same for this experiment except the hyperparameters are directly proportional to the training set size -- a very simple rule.  The learning rate step size is adjusted as \{100, 300, 500\} and training iterations of \{500, 1200, 2000\} for the training set size of \{25, 75, 125\} respectively.  Similar scaling was performed for the \{15, 35\} training sizes as well.   

Figure \ref{fig_ft_plots} shows Spearman's rank correlation performance at every hundred training iterations on the test set for each of the different training set sizes for all the actions. Finetuning from multi-action pretrained model is shown by blue-colored curves while red curves represent the result of finetuning from scratch (\ie random initialization). Overall, the blue curve is above the red in most situations.  Further, the all-action blue curve quickly reaches a high value after only a few tens of iterations.  The random-initialization red curve takes longer to gradually update weights and climb from a low starting value.  Table \ref{tab_ri_5vs1} provides the the best performance during training where we see that all-action is better in 16 of the 18 cases.  

Note that even in the case of Gymvault and Skiing which seemed to have poor all-action initialization (from the zero-shot AQA experiment in Table \ref{tab_zeroshot_aqa}) the performance quickly climbs during fine-tuning.  Deep networks don't typically have smooth error surfaces. So, it could be the case that there are valleys near good solution in cases of Gymvault and Skiing; and during finetuning, a good solution was reached because relatively higher initial learning rate of 0.001 helped in escaping local minima.  Individually setting hyperparameters for each individual action may further enhance performance but is outside the scope of this work.  Instead the emphasis here is on the ability to learn generalizable quality concepts by pre-training on multiple actions and providing a better initialization for fine-tuning.  
\subsection{Discussion}
The key outcome of this experimental evaluation was that learning AQA models by learning from multiple actions is advisable.  The experiments showed that there are quality concepts shared among actions that can be learned.  In addition, another advantage of using the multi-action training procedure is enhanced scalability.  We also followed a simple rule to set the hyperparameters. Moreover, pre-training with datapoints from all actions provides a better initialization point for faster convergence with the need for fewer training samples when fine-tuning for a new action.  One experiment that was not performed is using the all-action model for pre-training and fine-tuning for a single action.  However, this is not advisable since the data has already been incorporated into the all-action model and may lead to over-fitting.  

Actions that we have considered in this work, although from different sports, have common action elements which is a condition necessary for the transfer to work. We have not yet explored transferring or sharing knowledge among actions from different domains. For \eg, our approach is not intended to share knowledge between a surgery task and a completely different task like gymnastic vault. To be able to apply our approach, tasks should be from same domains. In the surgery skills domain, one should consider sharing knowledge between tasks like Knot-tying, Needle-passing and Suturing. We are looking forward to exploring cross-domain knowledge sharing in future works.
%
%
\section{Conclusion}
\label{conclusion}
This work demonstrates that like many other computer vision tasks, such as object classification or action recognition, action quality assessment (AQA) can benefit from knowledge transfer/sharing by training a shared model across samples from multiple actions. We experimentally demonstrated this on our newly introduced dataset, AQA-7.  The experiments showed: 1) that by considering multiple actions, limited per-action data is better leveraged for improved per action performance (leveraging data more efficiently), 2) multi-action pre-training provides better initialization for novel actions, hinting at an underlying consistency in the notion of quality in actions (demonstrating that all-action model is generalizable than single-action models). The results hint at the potential to extend AQA beyond sports scenarios into more general actions.  
\paragraph{Acknowledgments:} We would like to thank Nicolas Gomez, Ryan Callier, and Cameron Carmendy for helping us with dataset collection.  
{\small
\bibliographystyle{ieee}
\bibliography{egbib}
}
\end{document}